\documentclass{article}

\usepackage{arxiv}

\usepackage[utf8]{inputenc} 
\usepackage[T1]{fontenc}    
\usepackage{hyperref}       
\usepackage{url}            
\usepackage{booktabs}       
\usepackage{amsfonts}       
\usepackage{nicefrac}       
\usepackage{microtype}      
\usepackage{lipsum}		
\usepackage{graphicx}
\usepackage{natbib}
\usepackage{doi}
\usepackage{enumitem}
\usepackage{amsmath}
 \usepackage{multirow}

\title{Sparse Multi-Stage Expert-Agent Routing for Complex Clinical Reasoning}

\author{
Sike Xiang \quad
Shuang Chen \quad
Qian Sun \quad
Jia Cheng \quad
Yusi Wei \quad
Amir Atapour-Abarghouei\thanks{Corresponding author}\\[3pt]
Department of Computer Science \\
Durham University \\[3pt]
\texttt{sike.xiang@durham.ac.uk} \quad
\texttt{shuang.chen@durham.ac.uk} \\
\texttt{qian.sun@durham.ac.uk} \quad
\texttt{jia.cheng@durham.ac.uk} \\
\texttt{yusi.wei@durham.ac.uk} \quad
\texttt{amir.atapour-abarghouei@durham.ac.uk}
}

\renewcommand{\shorttitle}{Sparse Multi-Stage Expert-Agent Routing}

\begin{document}
\maketitle

\begin{abstract}
	Complex clinical reasoning requires models to update diagnostic hypotheses as new evidence emerges and to coordinate different medical specialities under limited consultation resources. Existing LLM-based clinical reasoning systems typically perform single-pass prediction or rely on fixed multi-agent workflows, making expert participation either static or unnecessarily exhaustive. We propose Sparse Multi-Stage Expert-Agent Routing, a language-based clinical reasoning framework that models diagnosis as a stage-wise routing process. Given progressively available clinical evidence derived from multiple modalities, the framework maintains an evolving case state and adaptively activates a sparse set of medical expert agents, supported by expert-specific memory across stages. To evaluate free-text diagnostic conclusions beyond surface similarity, we further introduce ClinFEScore, a fact-aware semantic evaluation protocol for clinical reasoning outputs. On reconstructed multi-stage cases from MAC and AgentClinic-NEJM, our framework reduces the average number of activated experts from 17.0 to 3.0 whilst maintaining strong fact-level diagnostic quality. On 200 real-world hospital MDT cases, ClinFEScore correlates strongly with clinician judgements (Spearman's $\rho=0.81$; Pearson's $r=0.87$), whilst our method achieves 91.5\% clinician-verified diagnostic accuracy with approximately five expert-agent/LLM calls per case. These results support sparse stage-wise coordination as an efficient and clinically relevant approach to LLM-based clinical reasoning.
\end{abstract}


\section{Introduction}
\label{sec:introduction}

Clinical diagnosis rarely begins with a complete case. Presenting symptoms and examination findings are usually available first, whilst laboratory results, imaging, pathology and follow-up observations arrive as the investigation proceeds \cite{gamborg2023clinical, HENDRIKS2024104267}. Clinicians revise their judgement as this evidence accumulates, often with input from different medical specialities \cite{35ea8fc14c8e4c62b466c5194bd99b88}. The specialists involved may also change as the diagnostic question becomes more specific \cite{BONACONSA2025155127, matsumoto2025factors}. Most LLM-based clinical reasoning systems do not model this process directly. They receive a case description once and produce a final answer, or they follow a fixed multi-agent workflow in which the same agents participate throughout \cite{hager2024evaluation, gaber2025evaluating, zhou2025largelanguagemodelsdisease}. In either case, expert participation cannot respond naturally to changes in the evidence.

\begin{figure*}[t]
  \centering
  \includegraphics[width=\linewidth]{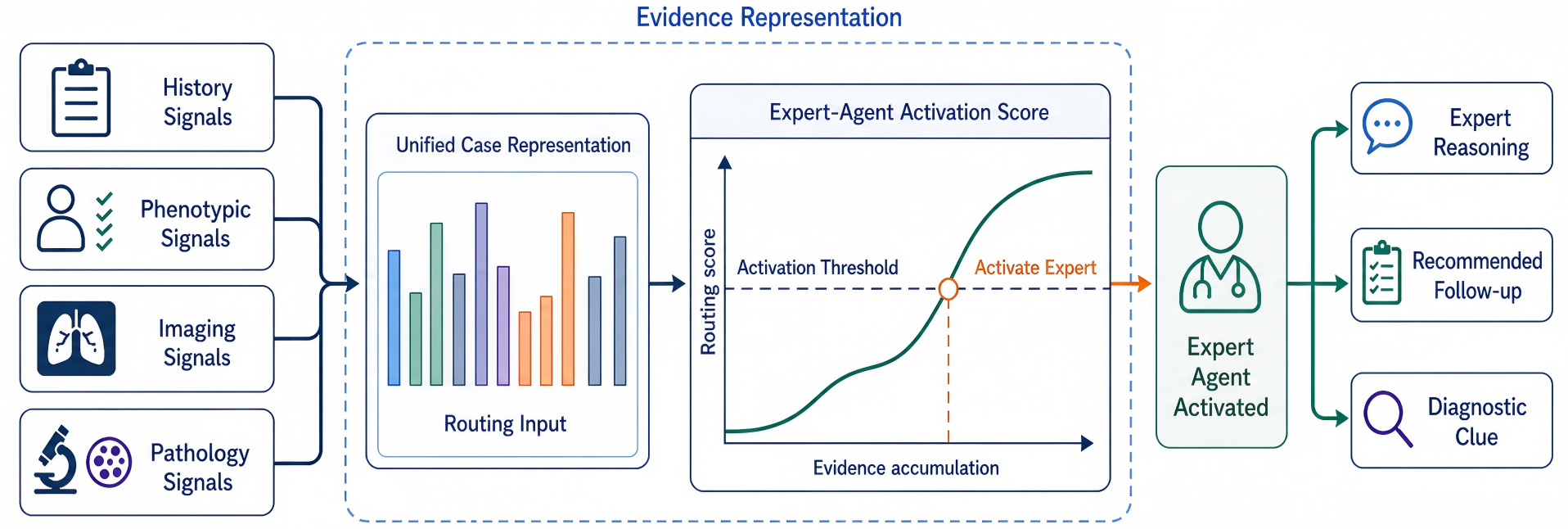}
  \caption{Illustration of the evidence-driven activation process for an individual expert agent within the sparse routing framework. Clinical evidence derived from multiple modalities is transformed into a unified case representation and an expert-agent activation score. When the score exceeds the adaptive threshold, the corresponding expert agent is invoked to produce stage-wise reasoning output.}
  \label{fig:1abc}\vspace{-0.5cm}
\end{figure*}

Modelling progressive clinical reasoning requires the system to preserve an evolving case state. New evidence should be able to revise an earlier hypothesis instead of being treated as an independent input \cite{tan2025assessment}. Expert participation must also remain conditional: whether an expert agent is useful depends on the evidence available at that stage and on what has already been inferred \cite{lamprell2024mdt}. Consultation is costly, so invoking every expert at every stage is neither clinically realistic nor computationally efficient \cite{alhammouri2024streamlining}. It can also introduce redundant opinions and expose each agent to information that is not relevant to its role. A useful routing mechanism must therefore decide which experts to invoke as the case develops, whilst retaining enough history to support diagnostic revision and an interpretable account of expert involvement \cite{elhaddad2024aidriven}.

Existing research addresses parts of this problem. Medical decision-support systems improve access to clinical knowledge and structured reasoning \cite{HENDRIKS2024104267, 10.1093/jamia/ocae209, NAFEES2023104143}. Mixture-of-experts models use conditional computation to activate subsets of latent model components \cite{Cai_2025, lo2025closerlookmixtureofexpertslarge, li2026motion}, whilst medical multi-agent systems organise reasoning around specialist roles or predefined consultation processes \cite{ke2024enhancingdiagnosticaccuracymultiagent, kim2024mdagentsadaptivecollaborationllms}. These approaches, however, make routing decisions at a single point or determine collaboration through a largely fixed workflow. They do not directly couple observable expert-agent participation to the gradual arrival of clinical evidence. As a result, it remains difficult to determine which expert should participate at a given stage, how earlier reasoning should influence that decision and what consultation cost the decision incurs.

We address this gap with Sparse Multi-Stage Expert-Agent Routing, a stage-aware framework for complex clinical reasoning. The framework treats expert participation as an evidence-dependent decision that is updated throughout the case. As shown in Fig.~\ref{fig:1abc}, the router encodes the current case state and textualised clinical evidence derived from multiple modalities, then assigns an activation score to each candidate expert agent. Adaptive thresholds and population-level inhibition keep the routing sparse, whilst a binary mask determines which expert agents are invoked for stage-wise reasoning. Their outputs update the case state before the next evidence stage arrives. Expert-specific memory preserves professional roles and relevant case context across stages, allowing later reasoning to build on earlier findings without repeatedly consulting the full expert pool. We also introduce ClinFEScore, a fact-aware semantic evaluation protocol that measures whether a free-text clinical conclusion is supported by and adequately covers the relevant clinical facts.

At the data level, we curate cases from MAC \cite{chen2025mac} and AgentClinic-NEJM \cite{schmidgall2024agentclinic} and organise them into sequential evidence blocks. Stage boundaries follow the order in which information becomes available during clinical investigation. Initial presentation, examination and preliminary laboratory findings appear first; imaging, pathology, genetic testing and follow-up evidence are introduced as the case develops. Each stage contains only the evidence available at that point, preventing later information from entering earlier reasoning. In addition to this controlled evaluation, we assess ClinFEScore and the routing framework on 200 real hospital multidisciplinary team (MDT) cases collected under ethics approval and data-use authorisation.

The experiments show that sparse routing reduces consultation cost without sacrificing the factual quality of the final conclusion. Relative to full multi-stage consultation, the complete framework reduces the average number of activated expert agents from 17.0 to 3.0 and lowers the simulated energy cost from 34.2 to 7.15, whilst maintaining comparable or better ClinFEScore results. On the 200 real-world MDT cases, three clinicians assessed diagnosis preservation, coverage of relevant evidence, fact omission and unsupported content. Their assessments show strong agreement, with a mean pairwise weighted $\kappa$ of 0.93. ClinFEScore correlates with these clinician judgements at Spearman's $\rho=0.81$ and Pearson's $r=0.87$. The framework also achieves 91.5\% clinician-verified diagnostic accuracy with approximately five expert-agent/LLM calls per case. Together, these findings support sparse stage-wise routing as a way to retain clinical content whilst avoiding unnecessary expert participation.

Our primary contributions are as follows:
\begin{enumerate}[label=(\roman*)]

    \item We propose Sparse Multi-Stage Expert-Agent Routing, a stage-aware framework that makes expert-agent activation responsive to evolving clinical evidence and prior reasoning state (Sec.~\ref{sec:method}).
    
    \item We introduce ClinFEScore, a fact-aware evaluation protocol for free-text clinical conclusions, and validate its agreement with clinician factual-consistency judgements on 200 real-world MDT cases (Sec.~\ref{sec:experiments:metrics} and App.~\ref{CFS}).
    
    \item We show through controlled experiments, real-world MDT evaluation and backbone ablations that sparse stage-wise routing preserves diagnostic quality whilst substantially reducing expert-agent invocation cost (Sec.~\ref{sec:experiments}).
\end{enumerate}

\section{Related Work}
\label{sec:relatedwork}
We consider prior work within the areas of complex clinical reasoning (Sec.~\ref{sec:rw:ccr}) and medical multi-agent reasoning with routing mechanisms (Sec.~\ref{sec:rw:mmar}).

\subsection{Complex Clinical Reasoning}
\label{sec:rw:ccr}

Complex clinical reasoning differs from static prediction tasks that map a single input directly to a final diagnosis. It requires clinicians to integrate heterogeneous evidence~\cite{Thampy2019ClinicalReasoning, Young2020ClinicalReasoningMapping}, including patient presentation, laboratory findings, imaging results and disease progression, whilst revising diagnostic judgement as new information becomes available~\cite{Andersson2019ClinicalReasoningEMS}. In complex cases, reasoning often extends beyond individual judgement to collaborative clinical reasoning, where multiple specialities jointly interpret evolving evidence and coordinate decision-making~\cite{Lee2024CollaborativeClinicalReasoning}. However, real-world multidisciplinary reasoning is constrained by limited expert availability, expanding medical knowledge and the need to integrate decision support into routine workflows~\cite{HENDRIKS2024104267}. Recent multimodal clinical-report benchmarks have also begun to move beyond free-text diagnosis by evaluating whether systems can reorganise heterogeneous clinical evidence into structured, safe and actionable outputs for patients~\cite{xiang2026checkup2action}. Building on this, our work formulates complex clinical reasoning as a sparse multi-stage expert-coordination problem, where evolving multimodal evidence updates the case state and determines which specialist agents should be activated under limited consultation resources.

\subsection{Medical Multi-Agent Reasoning}
\label{sec:rw:mmar}

Recent work has explored multi-agent paradigms for clinical reasoning, including diagnosis and simulated consultation~\cite{schmidgall2024agentclinic, tang-etal-2024-medagents}, alongside broader multi-agent collaboration frameworks that support role-specialised reasoning~\cite{wu2023autogen, park2023generative}. By decomposing reasoning into specialised roles, these systems can improve robustness and reduce some limitations of single-model inference. In parallel, mixture-of-experts and routing-based methods introduce conditional computation by selectively activating experts according to the input, often improving model capacity, specialisation or computational efficiency~\cite{Cai_2025,lo2025closerlookmixtureofexpertslarge}. The key limitation is therefore not simply the absence of multiple experts, but the absence of an explicit control mechanism for expert participation. Existing medical multi-agent systems typically define collaboration at the level of dialogue structure or agent roles, whereas routing-based models usually operate over latent representations rather than observable specialist consultations. This leaves the consultation process weakly coupled to clinical evidence progression, which makes it difficult to analyse which expert was invoked, at which stage and at what resource cost. Our work addresses this by making expert-agent activation itself the object of modelling, which leads to a sparse and stage-conditioned routing decision that links evolving case evidence to interpretable specialist participation.

\section{Method}
\label{sec:method}

\begin{figure*}[htbp]
    \centering
    \includegraphics[width=\textwidth]{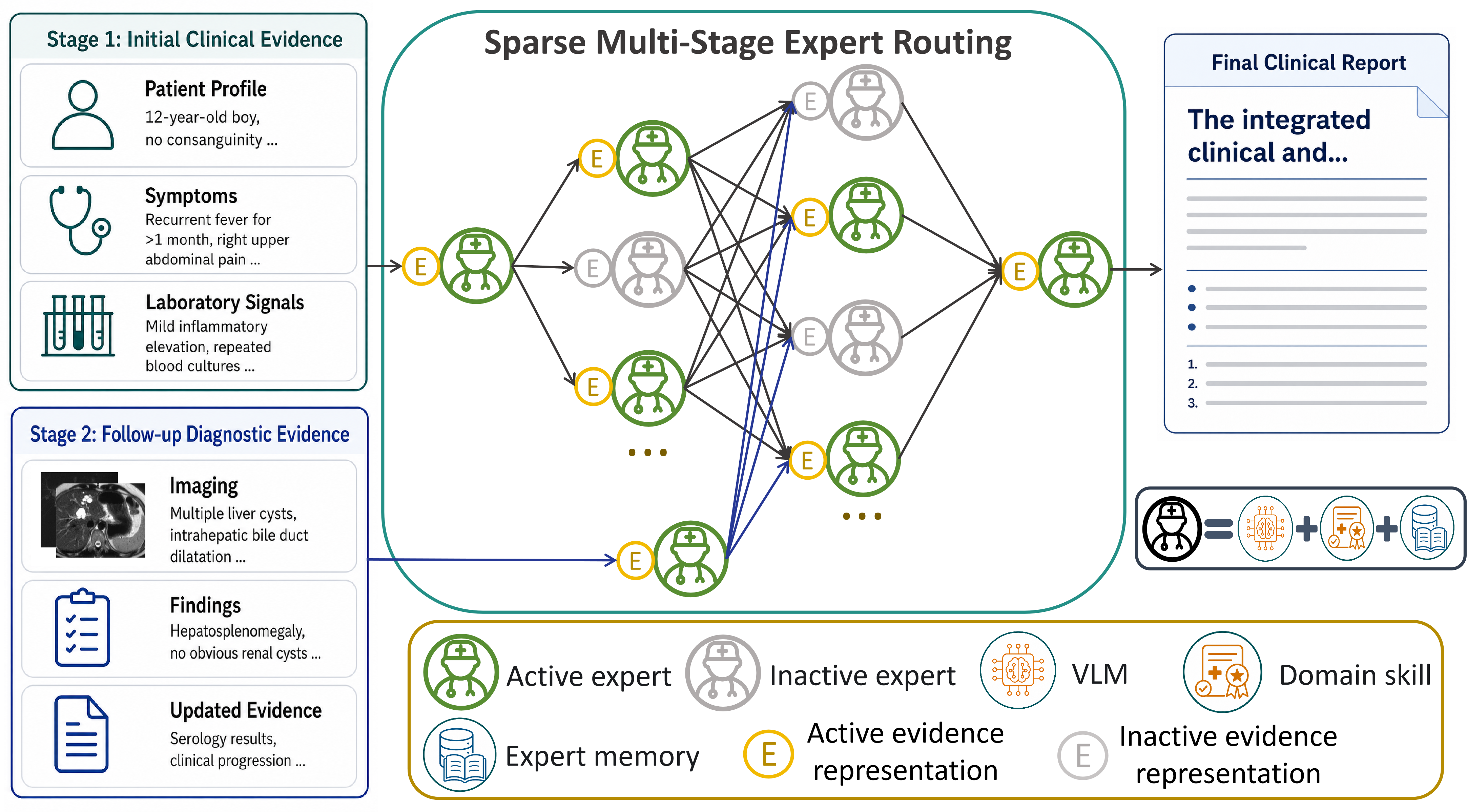}
    \caption{Overview of the proposed sparse multi-stage expert routing framework for complex clinical reasoning. Multimodal clinical evidence is organised into stage-wise evidence representations and routed through the activation mechanism in Fig.~\ref{fig:1abc}. The router selectively activates medical expert agents across stages, whose outputs are aggregated into a final clinical report. The clinical content shown above is for illustrative purposes only.}
    \label{fig:overall}\vspace{-0.5cm}
\end{figure*}

As shown in Fig.~\ref{fig:overall}, we propose a sparse multi-stage expert routing framework for complex clinical reasoning. The system receives multimodal clinical inputs, including patient information, presenting symptoms, physical examination findings, laboratory results, imaging evidence and follow-up diagnostic updates. A front-end clinical information layer organises these heterogeneous inputs into structured stage-wise evidence representations. At each reasoning stage, the routing module uses the current case state and evidence representation to selectively activate relevant medical expert agents, whose outputs contribute to stage-wise diagnostic reasoning. When additional evidence becomes available, the system combines the previous intermediate state with the newly observed evidence and updates the expert-agent activation pattern accordingly. Finally, stage-wise outputs are aggregated to produce the final diagnostic conclusion and an interpretable reasoning trajectory.

\subsection{Sparse Multi-Stage Expert-Agent Routing}
\label{sec:method:mssner}

The stage-wise routing process is defined as follows. Let the raw case input be denoted by:
\begin{equation}
x=\{x^{(1)},x^{(2)},\dots,x^{(K)}\},
\end{equation}
and let the initial unified case representation be:
\begin{equation}
h^{(0)} = f_{\mathrm{prep}}(x^{(1)}),
\end{equation}
where \(x^{(k)}\) represents the clinical evidence observed at stage \(k\) and \(f_{\mathrm{prep}}(\cdot)\) denotes the preprocessing and encoding function that maps the initial evidence into a unified case representation. This representation is then passed to the first-stage expert-agent layer. At stage \(s\), suppose there are \(N_s\) candidate expert agents. Given the previous state \(h^{(s-1)}\), the routing input for expert agent \(i\) is defined as:
\begin{equation}
I_i^{(s)} = W_i^{(s)} h^{(s-1)} + b_i^{(s)},
\end{equation}
and the internal routing state is updated by:
\begin{equation}
u_{i,t}^{(s)} = \lambda u_{i,t-1}^{(s)} + I_i^{(s)},
\end{equation}
where \(u_{i,t}^{(s)}\) is the internal routing state at internal step \(t\) and \(\lambda \in (0,1)\) is a decay coefficient controlling the retention and attenuation of historical state. Rather than performing static one-shot gating, the model updates routing states over a sequence of internal steps \(t=1,\dots,T_s\), allowing expert-agent activation to emerge progressively. To prevent persistent over-activation of the same expert agents, we further introduce an adaptive threshold mechanism:
\begin{equation}
\theta_{i,t}^{(s)} = \theta_0 + \alpha a_{i,t}^{(s)}, \quad a_{i,t}^{(s)} = \beta a_{i,t-1}^{(s)} + z_{i,t-1}^{(s)}, \quad z_{i,t}^{(s)} = \sigma\!\left(\gamma\left(u_{i,t}^{(s)}-\theta_{i,t}^{(s)}\right)\right).
\end{equation}
where \(\theta_0\) is the base threshold, \(\alpha\) is the adaptation coefficient, \(a_{i,t}^{(s)}\) is the adaptation state, \(\beta\) is a decay factor and \(z_{i,t}^{(s)}\) is the continuous expert-agent activation score. This mechanism discourages repeated over-activation of the same expert agent whilst keeping the continuous activation score differentiable for optimisation. After activation, we apply a soft attenuation to the internal state and compute the average activation intensity over the internal update window:
\begin{equation}
u_{i,t}^{(s)} \leftarrow u_{i,t}^{(s)} - z_{i,t}^{(s)} \theta_{i,t}^{(s)},
\end{equation}
\begin{equation}
\bar{z}_i^{(s)} = \frac{1}{T_s} \sum_{t=1}^{T_s} z_{i,t}^{(s)}.
\end{equation}

This yields a continuous estimate of stage-specific expert-agent activity. To avoid indiscriminate co-activation of all expert agents, we further impose a population-level inhibition mechanism coupled with target activation-rate control:
\begin{equation}
r^{(s)} = \frac{1}{N_s} \sum_{i=1}^{N_s} \bar{z}_i^{(s)},
\end{equation}
\begin{equation}
\tilde{\theta}_i^{(s)} = \theta_{\mathrm{sel}} + \eta \max(0, r^{(s)} - \rho),
\end{equation}
\begin{equation}
m_i^{(s)} = \mathbb{I}\!\left(\bar{z}_i^{(s)} > \tilde{\theta}_i^{(s)}\right),
\end{equation}
where \(r^{(s)}\) is the mean activity of stage \(s\), \(\theta_{\mathrm{sel}}\) is the base selection threshold, \(\rho\) is the target mean activation rate, \(\eta\) controls lateral inhibition when population activity becomes excessive and \(m_i^{(s)}\) is the final activation mask of expert agent \(i\). The continuous scores provide trainable routing signals, whereas the binary mask determines which expert agents are activated and invoked for stage-wise reasoning. For the activated expert agents, let \(e_i^{(s)}\) denote the stage-specific output of expert agent \(i\). We aggregate expert-agent responses into a stage representation and update the case state through:
\begin{equation}
g^{(s)} = \sum_{i=1}^{N_s} m_i^{(s)} \phi_i\!\left(e_i^{(s)}\right),
\end{equation}
\begin{equation}
h^{(s)} = f_{\mathrm{upd}}\!\left(h^{(s-1)}, g^{(s)}\right),
\end{equation}
where \(\phi_i(\cdot)\) denotes the expert-agent output projection function. Hence, the output of each stage is not treated as a final decision, but as an intermediate reasoning state that can influence subsequent stages. When new evidence \(x^{(s+1)}\) becomes available, the front-end organisation module encodes it as:
\begin{equation}
c^{(s+1)} = f_{\mathrm{prep}}(x^{(s+1)}),
\end{equation}
and the updated cross-stage state is:
\begin{equation}
\tilde{h}^{(s)} = [h^{(s)}; c^{(s+1)}],
\end{equation}
where \([\cdot;\cdot]\) denotes concatenation. For \(s=1,\dots,K-1\), the resulting cross-stage state is used as the routing input for the next stage:
\begin{equation}
I_i^{(s+1)} = W_i^{(s+1)}\tilde{h}^{(s)} + b_i^{(s+1)}.
\end{equation}
In this way, later stages do not simply repeat the first-stage computation, but incrementally incorporate new multimodal evidence on top of prior expert-agent reasoning. Finally, the outputs from all stages are integrated to produce the overall diagnostic prediction:
\begin{equation}
y = f_{\mathrm{out}}\!\left(g^{(1)}, g^{(2)}, \dots, g^{(K)}\right),
\end{equation}
where \(y\) denotes the final diagnostic conclusion. The resulting framework forms a stage-wise reasoning pipeline that spans heterogeneous input organisation, state construction, sparse expert-agent activation, cross-stage updating and final diagnostic aggregation.

\subsection{Loss Function}
\label{lf}

During training, we optimise the routing framework using a joint objective that combines final diagnostic consistency with routing regularisation. The retained losses include the final diagnostic consistency loss \(\mathcal{L}_{\mathrm{final}}\), the expert-agent invocation energy loss \(\mathcal{L}_{\mathrm{energy}}\), the activation entropy regularisation term \(\mathcal{L}_{\mathrm{entropy}}\) and the average activation-rate constraint \(\mathcal{L}_{\mathrm{rate}}\). The overall objective is defined as:
\begin{equation}
\mathcal{L} = \lambda_{\mathrm{final}} \mathcal{L}_{\mathrm{final}} + \lambda_{\mathrm{energy}} \mathcal{L}_{\mathrm{energy}} + \lambda_{\mathrm{entropy}} \mathcal{L}_{\mathrm{entropy}} + \lambda_{\mathrm{rate}} \mathcal{L}_{\mathrm{rate}}.
\end{equation}

The final diagnostic consistency loss is defined as:
\begin{equation}
\mathcal{L}_{\mathrm{final}} = 1 - R,
\end{equation}
where \(R\) denotes the semantic consistency score between the final diagnostic output and the reference answer. The expert-agent invocation energy loss is defined as:
\begin{equation}
\mathcal{L}_{\mathrm{energy}} = \sum_{a \in S_1} c(a) + \sum_{a \in S_2} c(a) + 0.5 \cdot \mathbb{I}(|S_1|>0) + 0.5 \cdot \mathbb{I}(|S_2|>0),
\end{equation}
where \(S_1\) and \(S_2\) denote the activated expert-agent sets in the first and second stages, respectively and \(c(a)\) denotes the invocation cost of expert agent \(a\). The two indicator terms impose a fixed overhead when at least one expert agent is activated in a stage, which reflects stage-level coordination cost. The activation entropy regularisation term is defined as:
\begin{equation}
\mathcal{L}_{\mathrm{entropy}} = \frac{1}{N_1}\sum_{i=1}^{N_1} H(s_{1,i}) + \frac{1}{N_2}\sum_{j=1}^{N_2} H(s_{2,j}),
\end{equation}
where:
\begin{equation}
H(p) = -p\log p - (1-p)\log(1-p),
\end{equation}
and \(s_{1,i}\) and \(s_{2,j}\) denote the activation intensities of the expert agents in the two stages. The average activation-rate constraint is defined as:
\begin{equation}
\mathcal{L}_{\mathrm{rate}} = \left|\frac{1}{N_1}\sum_{i=1}^{N_1}s_{1,i} - \rho_1\right| + \left|\frac{1}{N_2}\sum_{j=1}^{N_2}s_{2,j} - \rho_2\right|.
\end{equation}
where \(\rho_1\) and \(\rho_2\) denote the target average activation rates for the two stages. In implementation, we set \(\lambda_{\mathrm{final}}=0.6\), \(\lambda_{\mathrm{energy}}=0.15\), \(\lambda_{\mathrm{entropy}}=0.05\) and \(\lambda_{\mathrm{rate}}=0.2\). This objective encourages the model to balance final diagnostic quality, sparse expert-agent utilisation and stable routing behaviour. Detailed domain skill and expert memory designs are provided in App.~\ref{sec:method:dsd} and App.~\ref{sec:method:emd}, respectively.

\section{Experiments}
\label{sec:experiments}

\subsection{Dataset}
\label{sec:experiments:dataset}

We conduct experiments on two complementary sources of complex clinical cases, MAC \cite{chen2025mac} and AgentClinic-NEJM \cite{schmidgall2024agentclinic}. Together, they cover diagnostic scenarios involving long-form clinical narratives, structured findings, follow-up updates and multimodal clinical evidence such as imaging, laboratory, pathology and genetic information. Since the original data are not organised for multi-stage expert routing, we reconstruct each case into stage-wise inputs. The first stage contains initial clinical evidence, including demographic information, presenting symptoms, physical examination findings and preliminary laboratory results, whereas later stages introduce newly available diagnostic evidence such as imaging findings, pathology results, genetic tests and follow-up updates. This reconstruction supports dynamic expert-agent coordination and enables the dataset to be used not only for final diagnosis prediction, but also for analysing stage-wise evidence accumulation and expert-agent routing behaviour in complex clinical reasoning.

\subsection{Setting}
\label{sec:experiments:setting}

\begin{table*}[t]
\centering
\Large

\resizebox{0.95\textwidth}{!}{
\begin{tabular}{c c c c c  c c c  c c}
\toprule
\multirow{2}{*}{\textbf{SA}} & \multirow{2}{*}{\textbf{ST}} & \multirow{2}{*}{\textbf{MT}} & \multirow{2}{*}{\textbf{S}} & \multirow{2}{*}{\textbf{M}} & \multicolumn{3}{c}{\textbf{ClinFEScore} $\uparrow$} & \multirow{2}{*}{\textbf{Average Activated Experts} $\downarrow$} & \multirow{2}{*}{\textbf{Energy} $\downarrow$} \\
\cmidrule(lr){6-8}
 &  &  &  &  & \textbf{Fact Similarity} & \textbf{Fact Precision} & \textbf{Fact Recall} &  &  \\
\midrule
\multicolumn{5}{c}{\textbf{Ground Truth}} 
& 0.9972$_{\pm 0.0049}$ 
& 0.9904$_{\pm 0.0079}$ 
& 0.9887$_{\pm 0.0076}$ 
& -- & -- \\
\midrule
\checkmark & --         & --         & --         & --         
& 0.9434$_{\pm 0.0050}$ 
& 0.9590$_{\pm 0.0066}$ 
& 0.9556$_{\pm 0.0068}$ 
& 1.0000$_{\pm 0.0000}$ 
& 1.2000$_{\pm 0.0000}$ \\

--         & \checkmark & --         & --         & --         
& 0.9537$_{\pm 0.0046}$ 
& 0.9691$_{\pm 0.0083}$ 
& 0.9676$_{\pm 0.0073}$ 
& 9.0000$_{\pm 0.0000}$ 
& 17.1000$_{\pm 0.0000}$ \\

--         & \checkmark & \checkmark & --         & --         
& 0.9845$_{\pm 0.0043}$ 
& 0.9706$_{\pm 0.0071}$ 
& 0.9690$_{\pm 0.0066}$ 
& 17.0000$_{\pm 0.0000}$ 
& 34.2000$_{\pm 0.0000}$ \\

--         & \checkmark & \checkmark & \checkmark & --         
& 0.9936$_{\pm 0.0050}$ 
& 0.9788$_{\pm 0.0068}$ 
& 0.9768$_{\pm 0.0056}$ 
& \textbf{3.0119$_{\pm 0.2437}$} 
& \textbf{7.1488$_{\pm 0.8047}$} \\

--         & \checkmark & \checkmark & \checkmark & \checkmark 
& \textbf{0.9940$_{\pm 0.0043}$} 
& \textbf{0.9792$_{\pm 0.0063}$} 
& \textbf{0.9780$_{\pm 0.0062}$} 
& \textbf{3.0119$_{\pm 0.2437}$} 
& \textbf{7.1488$_{\pm 0.8047}$} \\
\bottomrule
\end{tabular}
}
\caption{Comparison of different system configurations, showing the effects of single-agent inference (SA), single-stage full expert-agent consultation (ST), multi-stage full expert-agent consultation (MT), sparse expert-agent routing (S) and memory augmentation (M). Ground Truth is reported as an upper-bound reference, and the best model results excluding Ground Truth are shown in \textbf{bold}. \texttt{gemini-3.1-pro-preview} is used as the backbone model.}
\label{tab:main_results}\vspace{-0.5cm}
\end{table*}

In our experiments, we use a unified two-stage expert-agent routing framework with a fixed expert-agent pool, including an initial triage agent and medical expert agents for general internal medicine, respiratory medicine, infectious disease, radiology, pathology, laboratory/microbiology, immunology/rheumatology and genetics/rare disease. For each case, a front-end clinical information organisation module parses the raw materials and converts the available multimodal clinical evidence into stage-wise textual representations. These representations are encoded using BiomedBERT \cite{10.1145/3458754} and passed to the two-stage routing module for expert-agent activation.

The MAC and AgentClinic-NEJM cases are split into training and validation sets with an \(8{:}2\) ratio. The routing module is trained for 20 epochs with a batch size of 1 using AdamW, and the learning rate of the routing layers is set to \(2\times10^{-4}\). Both Stage 1 and Stage 2 use four internal routing update steps. The training objective combines final diagnostic consistency, activation-rate regularisation, energy constraints and activation entropy regularisation to jointly optimise diagnostic quality, routing sparsity and overall stability. Unless otherwise specified, expert memory, short-term memory and long-term memory are enabled to support expert-agent reasoning, cross-stage state propagation and cross-case experience reuse.

\subsection{Evaluation Metrics}
\label{sec:experiments:metrics}

We evaluate model performance from two perspectives: diagnostic quality and routing efficiency. For diagnostic quality, we introduce \textbf{ClinFEScore}, a fact-aware free-text clinical semantic evaluator inspired by the fact-aware embedding-based framework of CXRFEScore \cite{messina2024extractingencodingleveraginglarge} and the fact extraction perspective of RadFact \cite{Bannur2024MAIRA2GR}. We adapt these ideas to complex multi-speciality clinical reasoning rather than chest radiograph reports alone. ClinFEScore first uses an LLM-based fact extractor to identify clinically important factual statements from both the generated conclusion and the reference answer and then encodes the extracted facts into a shared biomedical semantic space for comparison. We report three sub-metrics: \textit{Fact Similarity}, which measures the overall semantic similarity between the extracted fact sets; \textit{Fact Precision}, which measures how well the reference answer supports the facts expressed in the generated conclusion; and \textit{Fact Recall}, which measures how well the generated conclusion covers the clinically relevant facts contained in the reference answer. The detailed implementation of ClinFEScore is presented in App.~\ref{CFS}. For routing efficiency, we report \textbf{Average Activated Expert Agents}, which measures the average number of expert agents invoked per case and \textbf{Energy} (App.~\ref{ES}), which quantifies the relative resource cost of expert-agent collaboration.

\subsection{Quantitative Results}
\label{sec:exp:quantitative}

Tab.~\ref{tab:main_results} presents the quantitative results under different system configurations. Ground Truth is not treated as directly comparable to model outputs, but is included as an upper-bound reference for interpreting the ClinFEScore metrics. It achieves the highest values on all three sub-metrics, with 0.9972, 0.9904 and 0.9887 for Fact Similarity, Fact Precision and Fact Recall, respectively. This suggests that the proposed evaluation framework assigns the strongest scores to clinically complete reference conclusions, rather than merely rewarding superficial textual similarity. Ground Truth therefore provides a useful reference point against which the remaining configurations can be interpreted.

The other configurations reveal the incremental contribution of the major system components. The simplest single-agent setting achieves 0.9434, 0.9590 and 0.9556, showing that a single expert agent is insufficient for complex clinical reasoning, although it activates only one expert agent on average and incurs the lowest energy cost of 1.2. Expanding the system to single-stage full expert-agent consultation improves the three scores to 0.9537, 0.9691 and 0.9676, but increases the average number of activated expert agents to 9.0 and the energy cost to 17.1. When multi-stage full expert-agent consultation is introduced, performance further rises to 0.9845, 0.9706 and 0.9690, suggesting that stage-wise evidence modelling improves clinical conclusion quality. However, this configuration is also the most expensive, activating 17.0 expert agents on average and reaching an energy cost of 34.2. Once sparse dynamic routing is introduced, the model maintains strong performance whilst sharply reducing cost as without memory augmentation, the three scores reach 0.9936, 0.9788 and 0.9768, whereas the average number of activated expert agents drops to 3.0119 and the energy cost to 7.1488. Finally, adding memory augmentation on top of sparse routing yields the best model performance excluding Ground Truth, with 0.9940, 0.9792 and 0.9780, whilst keeping both activated expert-agent count and energy unchanged.

\begin{table*}[t]
\centering
\small

\resizebox{0.75\linewidth}{!}{
\begin{tabular}{c c c c}
\toprule
\multirow{2}{*}{\textbf{Backbone Model}} & \multicolumn{3}{c}{\textbf{ClinFEScore} $\uparrow$} \\
\cmidrule(lr){2-4}
& \textbf{Fact Similarity} & \textbf{Fact Precision} & \textbf{Fact Recall} \\
\midrule
Grok-4.20
& 0.9822$_{\pm 0.0078}$ 
& 0.9649$_{\pm 0.0061}$ 
& 0.9623$_{\pm 0.0058}$ \\

Claude-Opus-4.7
& 0.9829$_{\pm 0.0079}$ 
& 0.9650$_{\pm 0.0061}$ 
& 0.9640$_{\pm 0.0059}$ \\

GPT-5.4
& 0.9839$_{\pm 0.0080}$ 
& 0.9741$_{\pm 0.0070}$ 
& 0.9724$_{\pm 0.0068}$ \\

Gemini-3.1-Pro-Preview
& \underline{0.9940$_{\pm 0.0043}$} 
& 0.9792$_{\pm 0.0063}$ 
& 0.9780$_{\pm 0.0062}$ \\

\midrule
Hulu-Med-4B
& 0.9939$_{\pm 0.0049}$ 
& \underline{0.9795$_{\pm 0.0069}$} 
& \underline{0.9783$_{\pm 0.0062}$} \\

MedGemma-4B
& \textbf{0.9941$_{\pm 0.0050}$} 
& \textbf{0.9796$_{\pm 0.0071}$} 
& \textbf{0.9786$_{\pm 0.0067}$} \\
\bottomrule
\end{tabular}
}
\caption{Comparison of different backbone models under the same routing and memory configuration. The upper block lists general-purpose LLMs, whilst the lower block lists medical-domain LLMs. The best results are shown in \textbf{bold} and the second-best results are \underline{underlined}.}
\label{tab:backbone_models}\vspace{-0.5cm}
\end{table*}

\subsection{Ablation Studies}
\label{sec:exp:ablation}

\subsubsection{Backbone Ablation}
\label{sec:exp:ablation:BA}

Tab.~\ref{tab:backbone_models} compares different backbone models under the same routing configuration and otherwise matched memory setting. The medical-domain models in the lower block are deployed locally rather than accessed through network-enabled APIs. As a result, the expert memory component that depends on external retrieval is unavailable in this setting, although short-term and role-specific memory mechanisms are still retained. MedGemma \cite{medgemma} provides two versions, but only the 4B variant is multimodal, whereas the 27B variant is text-only. We therefore use \texttt{MedGemma-4B}. To keep model scale roughly aligned, we further select \texttt{Hulu-Med-4B} \cite{hulumed} as the corresponding 4B medical baseline.

The results show that the performance gap among mainstream backbone models is relatively small, which suggests that the proposed sparse multi-stage expert-agent framework is not overly dependent on a single backbone model. Although the two medical-domain models are smaller and operate without retrieval-based expert memory, their stronger domain alignment allows them to remain competitive, with \texttt{MedGemma-4B} achieving the best overall results at 0.9941, 0.9796 and 0.9786. This pattern suggests that ClinFEScore is sensitive to fact-level clinical consistency, whilst being less affected by superficial wording differences across backbone models. 

\subsubsection{Routing Layers Ablation}
\label{RLA}

As shown in Tab.~\ref{tab:layer_depth} in App.~\ref{RLA_f}, we compare the original two-layer routing structure with an additional four-layer configuration to examine the effect of routing depth. The current dataset is more naturally aligned with a two-stage reasoning process, corresponding to the initial integration of clinical evidence and the subsequent incorporation of key diagnostic findings. The four-layer configuration is therefore not intended to represent four naturally occurring clinical stages, but is introduced as a routing-depth ablation.

The results show that the four-layer structure does not provide a substantive gain on the three ClinFEScore metrics. Its \textit{Fact Similarity}, \textit{Fact Precision} and \textit{Fact Recall} are 0.9940, 0.9792 and 0.9780, respectively, differing only marginally from the two-layer results of 0.9941, 0.9798 and 0.9785. At the same time, the four-layer structure increases the average number of activated expert agents and the energy cost to 4.0000 and 9.4000, compared with 3.0119 and 7.1488 for the two-layer setting. Further inspection of the activation dynamics suggests a clear pattern collapse. The first two layers almost always select \texttt{general\_internal}, whilst the last two layers almost always select \texttt{radiology}. This indicates that, under the current data setting, increasing routing depth does not lead to richer expert-agent collaboration, but instead encourages the model to converge towards a fixed template-like routing strategy. These findings suggest that the two-layer structure is more appropriate for the reasoning logic of the current dataset.

\subsection{Real-World MDT Evaluation}
\label{sec:exp:real_mdt}

To assess ClinFEScore and sparse expert-agent routing in a clinical setting, we conduct an additional evaluation on 200 real hospital multidisciplinary team (MDT) cases. The study received ethics approval and data-use authorisation, and all cases underwent strict de-identification and clinical screening. The cases retain the order in which clinical evidence, speciality input and diagnostic discussion became available in practice. Three clinicians independently assessed whether each generated conclusion preserved the correct diagnosis, covered the relevant clinical evidence, omitted important facts or introduced unsupported information. Their assessments show strong agreement, with a mean pairwise weighted $\kappa$ of 0.93. ClinFEScore also correlates strongly with clinician factual-consistency judgements, achieving Spearman's $\rho=0.81$ and Pearson's $r=0.87$. These results indicate that ClinFEScore reflects both the ranking and score variation of clinician assessments in the real-world MDT setting.

We further compare our framework with MedAgents \cite{tang-etal-2024-medagents} and MDAgents \cite{kim2024mdagentsadaptivecollaborationllms} using the same 200 MDT cases, backbone model and final-diagnosis protocol. Because the frameworks differ in their output formats and internal evaluation measures, we use clinician-verified diagnostic accuracy as the common quality measure and report the average number of expert-agent or LLM calls per case as the consultation cost. Our method correctly diagnoses 183 of the 200 cases, corresponding to an accuracy of 91.5\%, whilst requiring approximately five calls per case. MDAgents correctly diagnoses 174 cases with approximately seven calls per case, whereas MedAgents correctly diagnoses 166 cases with approximately seventeen calls per case. These results show that sparse multi-stage expert-agent routing maintains strong diagnostic performance whilst avoiding unnecessary expert-agent invocation.

\section{Conclusion}
\label{sec:concl}

In this work, we propose Sparse Multi-Stage Expert-Agent Routing for complex multimodal clinical reasoning. By combining stage-wise clinical information organisation (Sec.~\ref{sec:method}), sparse expert-agent activation and collaborative reasoning (Sec.~\ref{sec:method:mssner}) and fact-aware semantic evaluation (Sec.~\ref{sec:experiments:metrics}), the framework supports structured diagnostic decision-making whilst balancing diagnostic quality and reasoning efficiency. We further introduce ClinFEScore (App.~\ref{CFS}), which assesses the semantic consistency and completeness of free-text clinical conclusions at the factual level. Controlled experiments show that multi-stage modelling, sparse routing and expert memory provide complementary benefits, reducing the average number of activated expert agents from 17.0 to approximately 3.0. The real-world MDT evaluation further shows that ClinFEScore correlates strongly with clinician factual-consistency judgements and that the framework achieves 91.5\% clinician-verified diagnostic accuracy with approximately five expert-agent or LLM calls per case (Sec.~\ref{sec:exp:real_mdt}). Results across different backbone models also suggest that these gains arise primarily from the proposed routing framework rather than from a single language model.

Future work will examine the framework through broader cross-institutional and prospective validation, with greater coverage of longitudinal evidence, speciality-specific activation patterns and variation in clinical workflows. We also plan to investigate stronger medically specialised backbone models, stricter factual-consistency constraints and more detailed measurements of token, latency and financial inference costs. These extensions will help determine how reliably sparse expert-agent routing transfers across clinical settings and where additional safeguards are required.

\bibliographystyle{unsrtnat}
\bibliography{references}  






\appendix

\section{Domain Skill Design}
\label{sec:method:dsd}

In this work, domain skills are implemented as role-specific professional prompts that provide stable prior knowledge for each expert agent. Each medical expert agent is assigned a set of skill descriptions aligned with its professional identity, specifying its responsibility scope, clinical focus, preferred evidence types and reasoning style. These skills are not learned through additional training, but are instantiated as structured prompt templates and provided to the backbone model together with the current case content and corresponding memory information when the expert agent is activated. This design helps different expert agents maintain stable and distinguishable professional perspectives, thereby supporting professional specialisation and role consistency in multi-expert collaboration.

\section{Expert Memory Design}
\label{sec:method:emd}

\begin{figure}[h]
  \centering
  \includegraphics[width=0.45\linewidth]{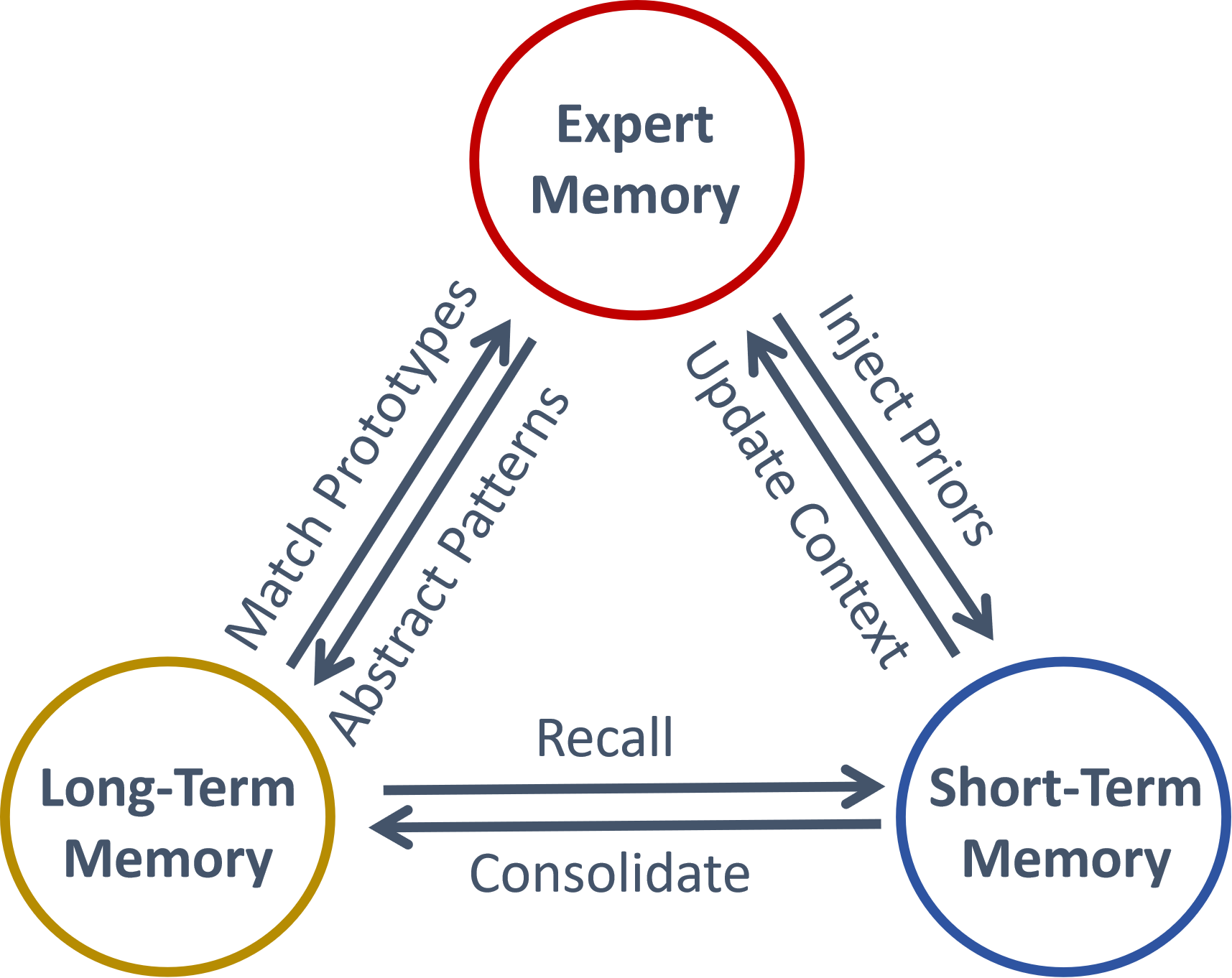}
  \caption{Expert memory structure. The memory module is organised as a triangular interaction among expert memory, short-term memory and long-term memory, supporting stable yet adaptive reasoning across stages.}
  \label{fig:memory}
\end{figure}

As shown in Fig.~\ref{fig:memory}, we organise the memory module as a stable triangular interaction among expert memory, short-term memory and long-term memory. Rather than treating each expert-agent invocation as an isolated response to the current input alone, this structure allows the system to jointly maintain domain-specific priors, case-specific evolving context and reusable cross-case experience. In particular, expert memory injects stable domain priors into the reasoning process, short-term memory preserves and updates the active case context across stages and long-term memory stores and recalls reusable experience patterns from prior cases. In this way, memory supports both stable expertise and adaptive reasoning across multiple stages of clinical decision-making.

Expert memory, denoted by \(M^{i}_{\mathrm{expert}}\), corresponds to the relatively stable memory associated with expert agent \(i\). It encodes the expert agent's domain role, reasoning scope, preferred evidence sources and high-priority diagnostic cues. In practice, this memory acts as a domain anchor, ensuring that each expert agent remains aligned with its intended speciality even as the clinical evidence evolves over stages. Short-term memory, denoted by \(M_{\mathrm{short}}^{(t)}\), stores the temporary working state of the current case at reasoning step \(t\), including extracted evidence, intermediate judgements, recognised salient features and contextual outputs exchanged among expert agents. This component maintains continuity within the current case, enabling subsequent stages to build on earlier reasoning rather than restarting from scratch. Long-term memory, denoted by \(M_{\mathrm{long}}\), stores cross-case experience accumulated over multiple cases, including representative patterns, reusable feature combinations and compacted diagnostic knowledge. Importantly, long-term memory need not preserve all cases at the same level of granularity: highly distinctive cases may retain richer traces, whereas ordinary cases may gradually be compressed into more concise summarised representations.

The short-term memory is updated dynamically as new expert-agent outputs and case states become available. At reasoning step \(t\), we write:
\[
M_{\mathrm{short}}^{(t)} = f_{\mathrm{short}}\!\left(M_{\mathrm{short}}^{(t-1)},\, h^{(t)},\, g^{(t)}\right),
\]
where \(h^{(t)}\) denotes the current case state and \(g^{(t)}\) denotes the aggregated expert-agent output at that stage. This update allows short-term memory to preserve the evolving case-specific context across stages.

Long-term memory is updated through a consolidation process that transfers valuable information from the current case into reusable cross-case knowledge:
\[
M_{\mathrm{long}} \leftarrow f_{\mathrm{long}}\!\left(M_{\mathrm{long}},\, M_{\mathrm{short}}^{(t)}\right).
\]
This process may preserve detailed traces for rare or representative cases whilst compressing more common cases into concise pattern-level summaries. Conversely, when a new case is being reasoned over, the system can retrieve relevant prior experience from long-term memory according to the current state:
\[
r^{(t)} = \mathrm{Retrieve}\!\left(M_{\mathrm{long}},\, h^{(t)}\right),
\]
where \(r^{(t)}\) denotes the recalled experience patterns most relevant to the current case.

Expert memory is then combined with short-term and long-term information to influence the final stage-specific expert-agent output. For expert agent \(i\), we define the memory-aware expert-agent representation as:
\[
\tilde{e}_i^{(t)} = f_{\mathrm{mem}}\!\left(e_i^{(t)},\, M^{i}_{\mathrm{expert}},\, M_{\mathrm{short}}^{(t)},\, r^{(t)}\right),
\]
where \(e_i^{(t)}\) is the original expert-agent output before memory fusion. This formulation reflects the intuition that expert-agent reasoning should not depend solely on the current input, but should also be grounded in stable domain priors, informed by the evolving context of the present case and supported by relevant historical prototypes.

\section{Energy Setting}
\label{ES}

To quantify the relative resource cost of expert-agent routing, we assign a fixed energy weight to each expert agent according to the typical diagnostic burden associated with its activation, using the initial triage agent as the baseline with energy 1.0. Relatively low-cost expert agents such as \texttt{general\_internal} and \texttt{respiratory} are assigned 1.2, whilst \texttt{infectious\_disease} and \texttt{immunology\_rheumatology} receive slightly higher weights of 1.3 and 1.4 due to their broader dependence on laboratory work-up and differential reasoning. \texttt{lab\_microbiology}, \texttt{radiology}, \texttt{pathology} and \texttt{genetics\_rare\_disease} are assigned progressively higher weights of 2.0, 2.5, 3.0 and 3.5, reflecting the increasing burden of specialised testing, imaging interpretation, histopathology and genetic or rare-disease evaluation. The final aggregation agent, \texttt{fusion\_chief}, is assigned a score of 1.0, as it primarily performs decision fusion rather than initiating additional high-cost examinations. These energy weights were defined in consultation with clinicians as a practical simulation-oriented reference for relative specialist burden, rather than as a direct estimate of real-world financial or operational cost. Under this setting, the total energy of a case is defined as the sum of the weights of all selected expert agents. It is incorporated into training as a regularisation term to encourage more efficient expert-agent utilisation whilst maintaining diagnostic performance.

\begin{table*}[t]
\centering
\small
\resizebox{\linewidth}{!}{
\begin{tabular}{c c c c c c}
\toprule
\multirow{2}{*}{\textbf{Routing Layers}} & \multicolumn{3}{c}{\textbf{ClinFEScore} $\uparrow$} & \multirow{2}{*}{\textbf{Average Activated Experts} $\downarrow$} & \multirow{2}{*}{\textbf{Energy} $\downarrow$} \\
\cmidrule(lr){2-4}
& \textbf{Fact Similarity} & \textbf{Fact Precision} & \textbf{Fact Recall} & & \\
\midrule
2
& \textbf{0.9941$_{\pm 0.0050}$}
& \textbf{0.9798$_{\pm 0.0070}$}
& \textbf{0.9785$_{\pm 0.0066}$}
& \textbf{3.0119$_{\pm 0.2437}$}
& \textbf{7.1488$_{\pm 0.8047}$} \\

4
& 0.9940$_{\pm 0.0049}$
& 0.9792$_{\pm 0.0070}$
& 0.9780$_{\pm 0.0065}$
& 4.0000$_{\pm 0.0000}$
& 9.4000$_{\pm 0.0000}$ \\
\bottomrule
\end{tabular}
}
\caption{Comparison between two-layer and four-layer routing structures under the same backbone model and evaluation setting. The best results are shown in \textbf{bold}. \texttt{Gemini-3.1-Pro-Preview} is used as the backbone model.}
\label{tab:layer_depth}
\end{table*}

\section{ClinFEScore}
\label{CFS}

ClinFEScore is computed through fact extraction and semantic comparison. Given a model-generated clinical conclusion \(y\) and its corresponding reference answer \(y^*\), the system first applies an LLM-based fact extractor to convert the two free-text inputs into two sets of atomic clinical facts:
\begin{equation}
\mathcal{F}(y)=\{f_1,\dots,f_m\}, \qquad
\mathcal{F}(y^*)=\{g_1,\dots,g_n\}.
\end{equation}

Here, each \(f_i\) or \(g_j\) denotes a concise and diagnostically meaningful fact unit, such as a key symptom, imaging finding, pathological clue, or diagnostic conclusion. A biomedical text encoder is then used to map each extracted fact into a shared semantic vector space, denoted by:
\begin{equation}
\mathbf{e}(f_i), \qquad \mathbf{e}(g_j).
\end{equation}

For any pair of facts, their semantic proximity is measured using cosine similarity:
\begin{equation}
s(f_i, g_j)=
\frac{\mathbf{e}(f_i)^\top \mathbf{e}(g_j)}
{\|\mathbf{e}(f_i)\|\,\|\mathbf{e}(g_j)\|}.
\end{equation}

On this basis, ClinFEScore reports three sub-metrics. First, \textit{Fact Precision} measures the extent to which the facts expressed in the generated conclusion are supported by the reference answer and is defined as the average best-match similarity from each predicted fact to the reference fact set:
\begin{equation}
\mathrm{Fact\ Precision}
=
\frac{1}{m}\sum_{i=1}^{m}
\max_{1\leq j\leq n} s(f_i, g_j).
\end{equation}

Second, \textit{Fact Recall} measures the extent to which the clinically relevant facts contained in the reference answer are covered by the generated conclusion and is defined as the average best-match similarity from each reference fact to the predicted fact set:
\begin{equation}
\mathrm{Fact\ Recall}
=
\frac{1}{n}\sum_{j=1}^{n}
\max_{1\leq i\leq m} s(f_i, g_j).
\end{equation}

Finally, \textit{Fact Similarity} measures the overall semantic similarity between the two fact sets as a whole. In our implementation, the predicted facts and reference facts are first concatenated into two global fact summaries:
\begin{equation}
\tilde{f}(y)=\mathrm{Concat}(f_1,\dots,f_m),
\end{equation}
\begin{equation}
\tilde{f}(y^*)=\mathrm{Concat}(g_1,\dots,g_n),
\end{equation}
which are then encoded by the same biomedical text encoder. The overall fact similarity is defined as:
\begin{equation}
\mathrm{Fact\ Similarity}
=
\frac{\mathbf{e}(\tilde{f}(y))^\top \mathbf{e}(\tilde{f}(y^*))}
{\|\mathbf{e}(\tilde{f}(y))\|\,\|\mathbf{e}(\tilde{f}(y^*))\|}.
\end{equation}

Conceptually, this design inherits the key ideas of both reference metrics. It follows CXRFEScore \cite{messina2024extractingencodingleveraginglarge} by first extracting facts and then comparing texts in a fact-aware semantic space, and reflects the RadFact \cite{Bannur2024MAIRA2GR} perspective by distinguishing factual support from factual coverage. As a result, ClinFEScore does not merely evaluate whether two free-text conclusions are superficially similar, but instead measures whether the generated conclusion expresses clinically correct, semantically supported and sufficiently complete factual content.

\section{Routing Layers Ablation}
\label{RLA_f}

Tab.~\ref{tab:layer_depth} provides the detailed routing-layer ablation results.

\end{document}